\documentclass[10pt,twocolumn,letterpaper]{article}

\usepackage[pagenumbers]{cvpr} 

\definecolor{cvprblue}{rgb}{0.21,0.49,0.74}
\usepackage[pagebackref,breaklinks,colorlinks,allcolors=cvprblue]{hyperref}

\usepackage{graphicx}
\usepackage{amsmath}
\usepackage{amssymb}
\usepackage{booktabs}
\usepackage[accsupp]{axessibility}

\usepackage{tikz}
\usepackage{pgfplots}
\pgfplotsset{compat=1.18}
\usepackage{pifont} 
\usepackage{capt-of}

\usepackage[capitalize]{cleveref}
\crefname{section}{Sec.}{Secs.}
\Crefname{section}{Section}{Sections}
\Crefname{table}{Table}{Tables}
\crefname{table}{Tab.}{Tabs.}

\def\paperID{15} 
\def\confName{CVsports}
\def\confYear{2026}

\title{Pixels or Positions? Benchmarking Modalities in Group Activity Recognition}

\author{
Drishya Karki\textsuperscript{1} \quad
Merey Ramazanova\textsuperscript{1} \quad
Anthony Cioppa\textsuperscript{2} \quad
Silvio Giancola\textsuperscript{1} \quad
Bernard Ghanem\textsuperscript{1} \\[2mm]
\textsuperscript{1}King Abdullah University of Science and Technology (KAUST), Saudi Arabia \\
\textsuperscript{2}University of Liège, Belgium \\
{\tt\small \{drishya.karki, merey.ramazanova, silvio.giancola, bernard.ghanem\}@kaust.edu.sa} \\
{\tt\small anthony.cioppa@uliege.be}
}

\begin{document}
\maketitle
\begin{abstract}
Group Activity Recognition (GAR) is well studied on the video modality for surveillance and indoor team sports (\eg, volleyball, basketball). Yet, other modalities such as agent positions and trajectories over time, \ie tracking, remain comparatively under-explored despite being compact, agent-centric signals that explicitly encode spatial interactions. Understanding whether pixel (video) or position (tracking) modalities leads to better group activity recognition is therefore important to drive further research on the topic. However, no standardized benchmark currently exists that aligns broadcast video and tracking data for the same group activities, leading to a lack of apples-to-apples comparison between these modalities for GAR. In this work, we introduce \emph{SoccerNet-GAR}, a novel multimodal dataset curated from the $64$ matches of the football World Cup 2022 by cleaning, synchronizing, and transforming raw broadcast video, player tracking, and event annotation sources into a unified, research-ready benchmark. Specifically, the broadcast videos and player tracking modalities for $87{,}939$ group activities are synchronized and annotated with $10$ categories. Furthermore, we define a unified evaluation protocol to benchmark two strong unimodal approaches: (i) competitive video-based classifiers and (ii) tracking-based classifiers leveraging graph neural networks. In particular, our novel role-aware graph architecture for tracking-based GAR directly encodes tactical structure through positional edges connecting players by their on-pitch roles. Our tracking baseline achieves $77.8\%$ balanced accuracy and $57.0\%$ macro F1 compared to $60.9\%$ balanced accuracy and $50.1\%$ macro F1 for the strongest video model, while training with $7 \times$ less GPU hours and $479 \times$ fewer parameters ($180$K \vs $86.3$M). This study provides new insights into the relative strengths of pixels and positions for group activity recognition in sports.
\end{abstract}

\begin{figure}[ht]
    \centering
    \includegraphics[width=1.0\linewidth]{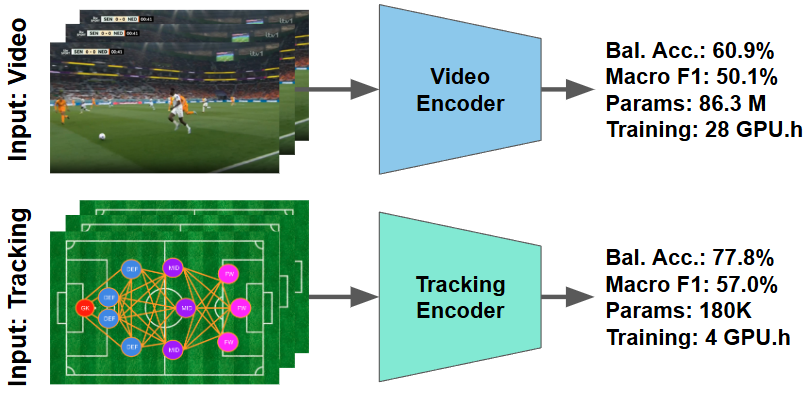}
    \caption{
    \textbf{Overview of Our Group Activity Recognition Benchmark.}
    Broadcast video and agent tracking modalities are processed through modality-specific backbones. The resulting representations are evaluated for group activity recognition.
}
    \label{fig:pulling}
\end{figure} 
\section{Introduction}
\label{sec:intro}
Group Activity Recognition (GAR) aims to identify collective behaviors that arise from multiple interacting agents rather than isolated individual actions. In team sports, this task is central to tactical analysis, automated coaching, and real-time decision support, where understanding both \emph{who} acts and \emph{how} players coordinate over space and time is essential. Effective GAR requires modeling individual dynamics together with their spatio-temporal relations at the scene level.
%
Most contemporary GAR models operate exclusively on RGB video, which offers appearance, context, and rich semantics. This choice also introduces practical and methodological limitations such as high computational cost, sensitivity to occlusions and editorial cuts, and a tendency to over-rely on appearance cues instead of explicitly modeling interactions. Structured representations based on agent positions and trajectories promise an alternative representation that is compact, agent-centric, and efficient to process. Because positions abstract away appearance, they may also generalize better across varying visual conditions. Progress, however, has been hampered by the absence of a standardized benchmark that aligns videos and agent positions per group activity, making apples-to-apples comparisons difficult. Furthermore, it remains unclear which edge connectivity and temporal pooling strategy are most appropriate for tracking based GAR.

In this paper, we address these gaps with a unified dataset and evaluation protocol for fair cross-modality evaluation. To do so, we introduce \emph{SoccerNet-GAR}, a novel multimodal dataset and benchmark built from the $64$ matches of the football World Cup 2022 by cleaning, merging, and aligning raw broadcast video, player tracking, and event annotation files from PFF FC (now Gradients Sports) into a synchronized, research-ready benchmark. The resulting dataset contains $87{,}939$ group activity instances, each pairing broadcast video and agent positions at the event level across $10$ football action classes. To support this new benchmark, we study two strong \emph{unimodal} approaches: a competitive video-based and a novel role-aware tracking-based classifier leveraging graph neural networks. On the 10-class group activity recognition task, our tracking baseline achieves \textbf{$77.8\%$} balanced accuracy and \textbf{$57.0\%$} macro F1 \vs \textbf{$60.9\%$} balanced accuracy and \textbf{$50.1\%$} macro F1 for the strongest video baseline, while training \textbf{$7 \times$} faster and using \textbf{$479\times$} fewer parameters.

\noindent\textbf{Contributions.} We summarize our contributions as follows:
\textbf{(i)}~We present \emph{SoccerNet-GAR}, a novel multimodal benchmark we curated from $64$ football World Cup matches by cleaning, merging, and temporally aligning separate raw files for broadcast video, player tracking, and event annotations. The resulting benchmark provides synchronized broadcast video and tracking data for $87{,}939$ events across $10$ classes, with match-level splits for fair, reproducible evaluation.
\textbf{(ii)}~We propose a \emph{novel role-aware, graph-based classifier for GAR} from tracking data, where tactical roles extracted from the source data define the graph connectivity, and conduct a comprehensive study of spatial connectivity and temporal aggregation design choices.
\textbf{(iii)}~We establish a \emph{pixels \vs positions} benchmark with identical protocols, report performance and efficiency comparisons, and distill practical design principles for tracking-based GAR.
The dataset, protocol, and code is available on \href{https://github.com/drishyakarki/pixels\_vs\_positions}{https://github.com/drishyakarki/pixels\_vs\_positions}.
\section{Related Work}
\label{sec: RelatedWork}
\subsection{Group Activity Recognition}
Group Activity Recognition (GAR) studies collective behaviors that arise from coordinated interactions among multiple agents, going beyond single-actor recognition \cite{choi2009they}. Early works relied on hand-crafted relational cues and hierarchical temporal models \cite{choi2009they,ibrahim2016hierarchical,bagautdinov2017socialscene}. The field then shifted toward deep learning approaches with graph-based architectures that explicitly encode inter-person dependencies \cite{fang2018learning,yan2018spatialtemporal,wang2020learning}, followed by transformer-based methods that model long-range spatial-temporal relationships through self-attention \cite{li2021groupformer,gavrilyuk2020actor}. Popular benchmarks include ``Collective Activity'' and ``Volleyball'', which focus on indoor team sports with controlled camera viewpoints \cite{choi2009they,ibrahim2016hierarchical}. 
Contemporary methods predominantly operate on RGB clips using 3D CNNs or vision transformers \cite{radford2021learning,tong2022videomae,wang2023videomae,simeoni2025dinov3}, often combined with person detection, region pooling, or attention over actor tokens to model interactions. Recent work has advanced relational modeling further: higher-order interactions through factor graphs~\cite{xie2024active} and bidirectional causality between human relations and human-object interactions~\cite{zhang2024bicausal}. While these
approaches achieve strong performance, they remain video-centric, and their relational reasoning is mediated through appearance features extracted from pixels. Several works explored structured alternatives other than raw videos. Pose- and skeleton-based approaches explicitly model spatial relationships between joints and have shown that structured representations can reduce reliance on background and scene bias while remaining competitive~\cite{zhou2022composer, li2024skeleton}. More broadly, tracking data from sports analytics systems offers a complementary modality by providing agent-centric trajectories and role metadata in a calibrated field coordinate frame, enabling explicit modeling of team structure and inter-agent geometry. Despite being compact, naturally encoding interactions, and robust to visual distribution shifts, tracking-based representations have received limited attention for GAR, and no existing benchmark evaluates these modalities under synchronized, comparable conditions. In our work, we address this gap by providing a fair, apples-to-apples evaluation of \emph{unimodal} pixels versus \emph{unimodal} positions under identical training protocols and matched event-level synchronization. 

\subsection{Sports Video \& Tracking Analytics}
Sports video understanding has advanced significantly with broadcast analysis spanning many sports including handball~\cite{Host2022AnOverview}, basketball~\cite{Istasse2023DeepSportradarv2}, and football~\cite{Giancola2024Deep}. The SoccerNet~\cite{giancola2018soccernet} and SoccerNet-v2~\cite{deliege2021soccernet} datasets established large-scale benchmarks for action spotting, expanding from $3$ to $17$ action classes with precise temporal annotations for every event during a game. Recent extensions include SoccerNet-v3~\cite{cioppa2022scaling} with multi-view player localization annotations, SoccerNet Ball Action Spotting~\cite{cioppa2024soccernet}, SoccerNet Game State Reconstruction~\cite{somers2024soccernet}, as well as many others~\cite{Leduc2024SoccerNetDepth, Cioppa2022SoccerNetTracking,Sarkhoosh2025Beyond,Gautam2024SoccerNetEchoes}.
In parallel to broadcast video analysis, structured tracking signals such as agent centroids and trajectories have been widely used in sports analytics for forecasting, pass modeling, and event detection~\cite{yeh2019forecasting, fernandez2019wide, cartas2022graphbased, Cioppa2021Camera}. Graph-based methods naturally capture coordinated play by representing players and the ball as nodes with relational edges \cite{stockl2021making, wang2023tacticaiaiassistantfootball, yeh2019diverse, anzer2022detection}. However, these studies typically address trajectory forecasting \cite{yeh2019diverse}, pass and defensive performance modeling \cite{stockl2021making, fernandez2019wide}, tactical pattern detection \cite{wang2023tacticaiaiassistantfootball} rather than multi-class group activity recognition under synchronized video-tracking alignment. The closest work is NETS \cite{hauri2022group}, which recognizes group activities from NBA tracking data using a Transformer-LSTM architecture, but it operates on a single modality without a matched video counterpart for cross-modality comparison. 
Our work addresses this gap by proposing the first synchronized dataset where broadcast video and tracking windows correspond to the same annotated group activities, and leveraging it to directly compare pixel-based and position-based representations under identical evaluation settings.
 
\subsection{Graph and Relational Modeling}
Graph neural networks enable relational reasoning for multi-agent scenes by propagating messages between agents \cite{kipf2016gcn,hamilton2017graphsage,brody2022attentivegraphattentionnetworks,xu2019powerfulgraphneuralnetworks}. When applying such models to multi-agent activity recognition, two design axes are particularly consequential: graph connectivity, which determines how agents exchange information, and temporal aggregation, which governs how frame-level representations are consolidated into a sequence-level prediction. EdgeConv \cite{wang2019dynamicedgeconv} enables learned, adaptive edges through dynamic neighborhood construction rather than fixed graphs. Training deeper GCNs has been challenging due to oversmoothing, but DeepGCNs \cite{li2019deepgcns} show that residual connections enable much deeper graph networks. Temporal modules range from efficient temporal convolutional networks \cite{lea2017temporalconvolutional} to recurrent \cite{graves2005framewise} or attention-based aggregators for long-range context \cite{gavrilyuk2020actor}.
While GNNs have been applied to sports tasks such as trajectory forecasting and action rating~\cite{yeh2019forecasting,ding2020graphattention,majeed2025realtime}, prior studies typically fix parts of the modeling pipeline or evaluate on limited data, making it difficult to extract modality-specific best practices. We complement this literature with a novel role-aware graph classifier for tracking-based GAR and conduct ablations across connectivity schemes (fixed, distance-based, positional edges) and temporal aggregation strategies (pooling, TCN, attention) to identify which design choices yield optimal accuracy-efficiency trade-offs.
\section{SoccerNet-GAR Dataset}
We present \textbf{SoccerNet-GAR}, a new multimodal dataset for Group Activity Recognition constructed from the $64$ matches of the football World Cup 2022 tournament. The raw data originate from PFF FC \cite{pff2022worldcupdataset}, which provides broadcast videos, player tracking data, and event annotations for each match. These modalities are distributed as separate files per match, not directly aligned for computer vision tasks. We transform this raw sports analytics data into a structured benchmark suitable for Group Activity Recognition by performing data cleaning, temporal synchronization, quality control, and window extraction around annotated events.

\subsection{Data Modalities}
Our dataset provides two input modalities: broadcast video and player tracking. Both originate from untrimmed full-game broadcast videos of the football World Cup 2022 matches, made available through PFF FC. 

\noindent\textbf{Video Modality.} 
PFF FC provides the edited broadcast footage at $720$p resolution. Each frame $I_t \in \mathbb{R}^{H \times W \times 3}$ is part of a temporal sequence $\mathbf{X}^V = \{I_t\}_{t=1}^{T}$ sampled within a temporal window. The video captures appearance cues, scene context, and visual motion patterns.

\noindent\textbf{Tracking Modality.} The tracking data consists of 2D player positions and 3D ball coordinates sampled at $30$ fps, automatically extracted from broadcast footage using computer vision tools and manually refined by PFF FC annotators. Player positions span $x \in [-60, 60]$m, $y \in [-42, 41]$m; ball positions include height $z \in [-8, 25]$m. Formally, the tracking input is represented as a temporal sequence of multi-agent states $\mathbf{X}^P = \{\mathcal{S}_t\}_{t=1}^{T}$, where each state $\mathcal{S}_t = \{s_t^1, \ldots, s_t^N\}$ describes $N$ entities (players and ball). Each entity state $s_t^i \in \mathbb{R}^D$ encodes spatial coordinates $(x, y) \in \mathbb{R}^2$ for players and $(x, y, z) \in \mathbb{R}^3$ for the ball, its identity, \ie one-hot encoding indicating entity type, and motion dynamics, \ie  displacement vectors $(\Delta x, \Delta y)$ capturing movement between consecutive frames. In addition, PFF FC provides positional role metadata for each player. We extract these roles and group them into four tactical categories (goalkeeper, defender, midfielder, forward) that we use as fixed role labels to define graph connectivity. These role categories are used to define structured connectivity patterns in our role-aware graph model. 

\subsection{Annotation \& Modality Alignment}
The PFF FC dataset~\cite{pff2022worldcupdataset} provides event annotations with precise timestamps for each action occurrence. Each event is labeled with one of $10$ football group activities (\textit{PASS}, \textit{TACKLE}, \textit{OUT}, \textit{HEADER}, \textit{THROW IN}, \textit{CROSS}, \textit{FREE KICK}, \textit{SHOT}, \textit{GOAL} and \textit{HIGH PASS}) and temporally marked at the moment of occurrence. Annotations were created by trained annotators and verified through quality control procedures by PFF FC using both video and tracking views.
To create our SoccerNet-GAR dataset, we align event annotations with both input modalities by merging them with tracking streams using UTC timestamps. We then apply three successive filters to ensure data quality: (1) we remove events where no tracking frame falls within a $10$\,ms tolerance of the event timestamp, discarding temporally misaligned entries, (2) we discard events that lack corresponding data in either modality, ensuring every retained sample has both video and tracking coverage, and (3) we resolve duplicate labels, where a single timestamp is annotated with more than one action class (\eg, a goal also labeled as a shot), by retaining only the most semantically specific label based on a predefined priority ordering. Together, these filters remove $6{,}346$ events ($6.8\%$ of the raw annotations), yielding a final dataset of $87{,}939$ annotated group activities. For each annotated event at timestamp $t_e$, we extract a temporal window of $4.5$ seconds centered at $t_e$ from both modalities, picking $T=16$ samples at $30$ fps with a $9$-frame interval (effectively sampled at $\approx 3.3$ fps over $4.5$ seconds), producing synchronized observations $\mathbf{X}^V = \{I_t\}_{t=1}^{T}$ and $\mathbf{X}^P = \{\mathcal{S}_t\}_{t=1}^{T}$, labeled with a single football group activity.

\begin{table}[t]
\centering
\resizebox{1.0\linewidth}{!}{%
\begin{tabular}{lccccccc}
\toprule
\textbf{Dataset} & \textbf{Year} & \textbf{Domain} & \textbf{Events} & \textbf{Cls.} & \textbf{Mod.} \\
\midrule
CAD~\cite{choi2009they} & 2009 & Pedestrian & 2,511 & 5 & V \\
Volleyball~\cite{ibrahim2016hierarchical} & 2016 & Volleyball & 4,830 & 8 & V \\
SoccerNet~\cite{giancola2018soccernet} & 2018 & Football & 6,637 & 3 & V \\
NBA~\cite{yan2020socialadaptivemoduleweaklysupervised} & 2020 & Basketball & 9,172 & 9 & V \\ 
SoccerNet-v2~\cite{deliege2021soccernet} & 2021 & Football & 110,458 & 17 & V \\
NETS~\cite{hauri2022group} & 2022 & Basketball & 61,053 & 3 & T \\
SoccerNet-BAS~\cite{cioppa2024soccernet} & 2024 & Football & 11,041 & 12 & V \\
Café~\cite{kim2023towards} & 2024 & Indoor & 10,297 & 6 & V \\
FIFAWC~\cite{pei2024fifawc} & 2024 & Football & 5,196 & 12 & V \\
\midrule
\textbf{SoccerNet-GAR} & \textbf{2026} & \textbf{Football} & \textbf{87,939} & \textbf{10} & \textbf{V + T} \\
\bottomrule
\end{tabular}%
}
\caption{
\textbf{Comparison of Group Activity Recognition Datasets.} 
'V' and 'T' indicate Video and Tracking, respectively.
SoccerNet-GAR is the second largest dataset in terms of events and the only one providing synchronized video and tracking modalities.
}
\label{tab:dataset_comparison}
\end{table}

\begin{figure}[t]
    \centering
    \includegraphics[width=\linewidth]{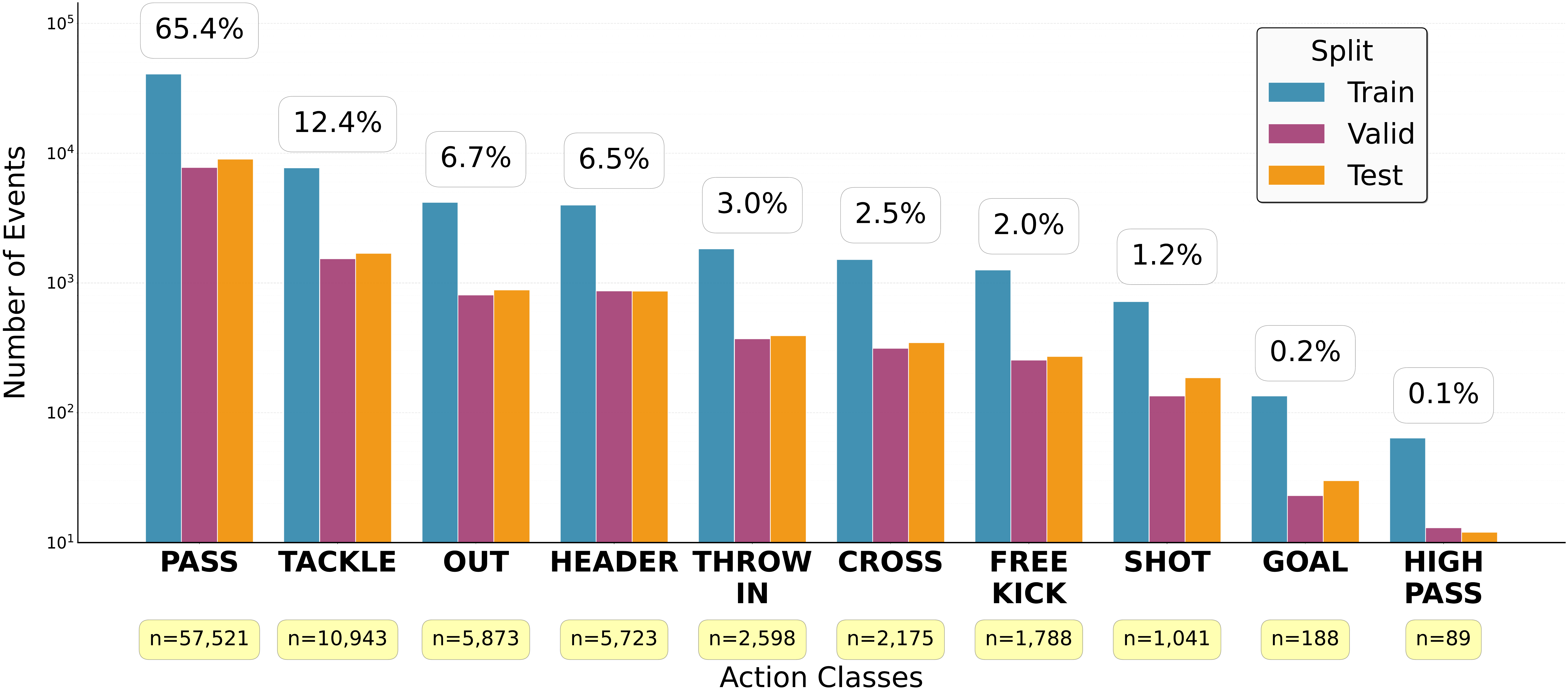}
    \caption{
    \textbf{Class Distribution Across Each Split.} 
    The dataset is heavily skewed toward \textit{PASS} ($65.4$\%) while \textit{GOAL} and \textit{HIGH PASS} are rare ($0.2$\% and $0.1$\%). The y-axis shows event counts; percentages denote proportions.
    }
    \label{fig:class_distribution}
\end{figure}

\subsection{Statistics}
The dataset contains $87{,}939$ annotated events across $10$ action classes, averaging $1{,}374$ events per match. To avoid leakage, data were split by match: $45$ training matches ($62{,}159$ events, $70.7$\%), $9$ validation matches ($12{,}091$ events, $13.7$\%), and $10$ test matches ($13{,}689$ events, $15.6$\%).

\noindent\textbf{Class Distribution.} 
Figure~\ref{fig:class_distribution} shows the distribution across classes and splits. Our dataset is highly imbalanced across classes, with \textit{PASS} events representing $65.4$\% of annotations while rare events like \textit{GOAL} and \textit{HIGH PASS} account for only $0.2$\% and $0.1$\%. This severe imbalance (646:1 ratio) reflects the natural distribution of football events and poses significant challenges for recognition models.

\noindent\textbf{Object Visibility and Tracking Quality.}
The tracking data maintain high completeness for players ($99.9$\% of $1{,}485{,}008$ frames contain all $11$ players per team) whereas ball is visible in $93.4$\% of frames, due to occlusions, camera transitions, and extreme zoom levels. At the event level, $85.9$\% of annotated events have complete ball tracking within their temporal window, while the remaining $14.1$\% contain at least one frame with missing ball data.

\subsection{Comparison with Existing Datasets}
Table~\ref{tab:dataset_comparison} compares our dataset with existing Group Activity Recognition benchmarks. 
With $87{,}939$ annotated events, SoccerNet-GAR is the second largest GAR dataset after SoccerNet-v2 ($110{,}458$ events), and substantially larger than other football datasets: $8\times$ larger than SoccerNet-BAS ($11{,}041$ events) and $16.9\times$ larger than FIFAWC ($5{,}196$ events). Our $10$ action classes overlap with the $12$ classes in SoccerNet-BAS, covering the same core football events. SoccerNet-GAR is the only dataset providing synchronized tracking and video modalities for the same action instances, enabling direct apples-to-apples comparison between tracking-based and video-based approaches. This dual-modality design distinguishes our dataset from prior work such as SoccerNet-v2~\cite{deliege2021soccernet} and SoccerNet-BAS~\cite{cioppa2024soccernet} which provide only video, while existing tracking datasets (\eg NETS~\cite{hauri2022group}) lack corresponding broadcast video.
\section{Methodology}
\subsection{Problem Formulation}
We formulate the Group Activity Recognition (GAR) problem as a multi-class classification task over temporal windows of length $T$, centered on annotated event timestamps. Given input from one modality, either video frames $\mathbf{X}^V = \{I_t\}_{t=1}^{T}$ or tracking data $\mathbf{X}^P = \{\mathcal{S}_t\}_{t=1}^{T}$, the task is to learn a function $f_\theta: \mathcal{X} \rightarrow \mathcal{Y}$ that predicts the activity label $y \in \mathcal{Y}$. 

\subsection{Recognition Framework}
Our recognition framework follows a standard backbone-neck-head architecture~\cite{benzakour2024oslactionspottingunifiedlibraryaction}, where modality-specific backbones extract spatial features, the neck temporally aggregates features to capture dynamics, and a classification head produces final predictions.

\noindent\textbf{Video Backbones.} For the video modality, we employ pretrained vision transformers~\cite{tong2022videomae,wang2023videomae} that encode spatiotemporal information from broadcast footage. Given an input clip $\mathbf{X}^V = \{I_t\}_{t=1}^{T}$, the backbone processes each frame $I_t$ through transformer layers with self-attention mechanisms, producing contextualized token embeddings. We extract frame-level representations $\mathbf{z}_t^V \in \mathbb{R}^{d_V}$ at each timestep $t$, yielding a temporal sequence $\{\mathbf{z}_t^V\}_{t=1}^{T}$ for subsequent temporal aggregation.

\noindent\textbf{Tracking Backbone.} For the tracking modality, we propose a graph-based architecture that explicitly models spatial relationships among entities at each timestep. At each frame $t$, we construct a graph $\mathcal{G}_t = (\mathcal{V}_t, \mathcal{E}_t)$ from the tracking data $\mathbf{X}^P = \{\mathcal{S}_t\}_{t=1}^{T}$, where nodes $\mathcal{V}_t = \{v_t^1, \ldots, v_t^N\}$ represent $N$ entities  and edges $\mathcal{E}_t$ encode their spatial interactions. Each node carries an 8-dimensional feature vector comprising its pitch coordinates $(x, y)$, instantaneous velocity $(\Delta x, \Delta y)$, ball height $z$ (zero for players), and a one-hot indicator of entity type (home, away, ball). A graph neural network processes these graphs to produce frame-level embeddings $\mathbf{z}_t^P \in \mathbb{R}^{d_P}$.

\noindent\textbf{Classification Task.}
Given frame-level representations $\{\mathbf{z}_t\}_{t=1}^{T}$, the temporal aggregation module synthesizes information across the observation window to produce a unified representation $\mathbf{\hat{z}} = f_{\text{temp}}(\mathbf{z}_1, \ldots, \mathbf{z}_T)$, where $f_{\text{temp}}$ is a temporal aggregation function. We evaluate various temporal aggregation strategies spanning from parameter-free pooling operations to learnable sequential models. A Multi-Layer Perceptron (MLP) maps the aggregated representation to class logits through a softmax activation. 
We follow a standard training protocol across modalities, minimizing a cross-entropy loss and randomly sampling $4,000$ actions per class and per epoch to address class imbalance.

\subsection{Graph Construction}
The edge set $\mathcal{E}_t$ defines which entities exchange information during message passing. We evaluate several connectivity schemes encoding different spatial interaction priors: \emph{no edges} (independent node processing), \emph{full connectivity} (all nodes linked), \emph{distance-based} (edges within radius $r$), \emph{KNN} ($k$ nearest neighbors), \emph{ball-centric} variants of distance and KNN (anchored on ball position), and our proposed \emph{positional} connectivity described below. These variants enable identifying which graph topologies effectively capture coordinated play.

\begin{figure}[t]
    \centering
    \includegraphics[width=1.0\linewidth]{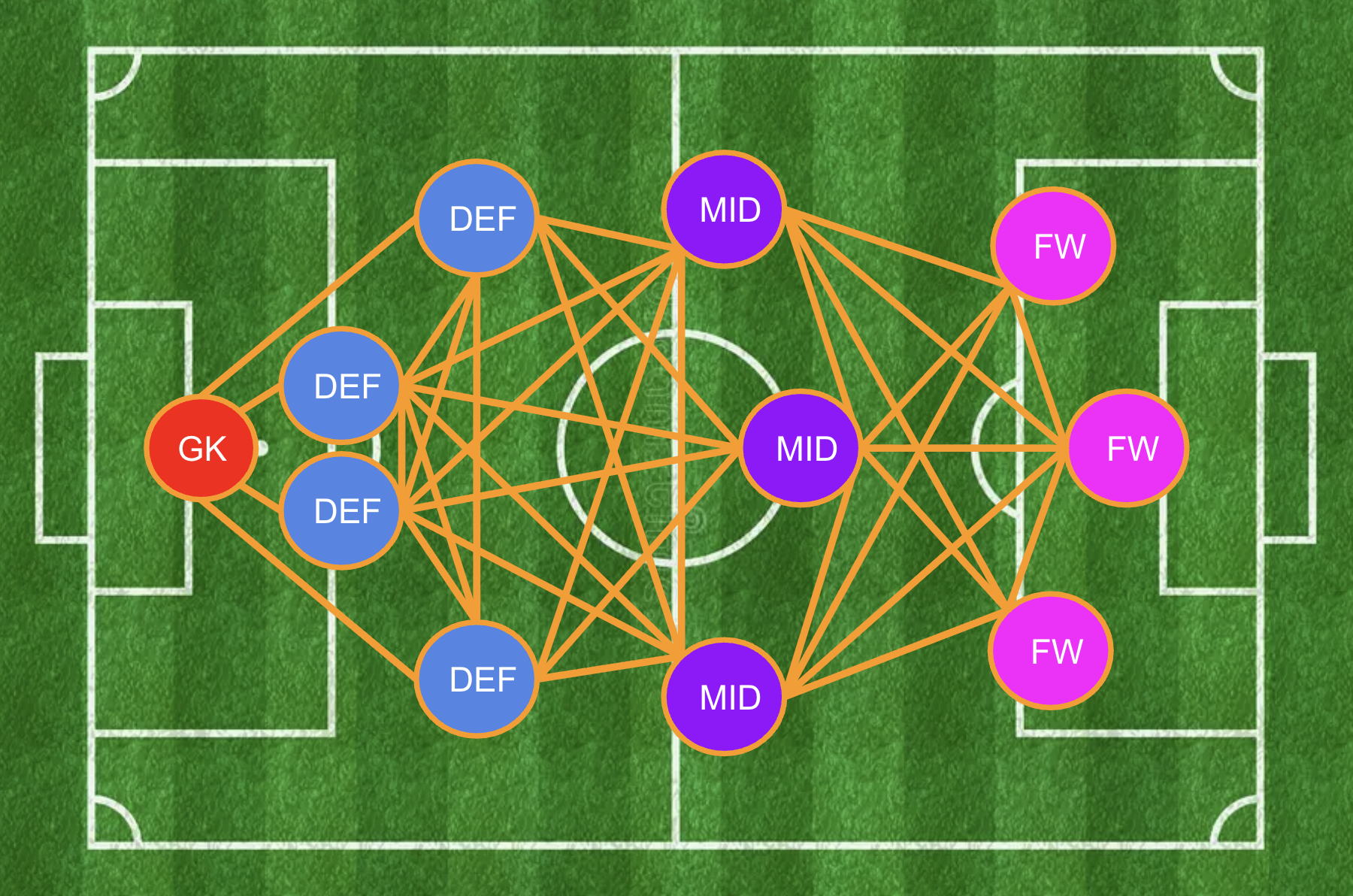}
    \caption{
    \textbf{Positional Graph Representation at a Single Frame.}
    Nodes represent players at their pitch coordinates, colored by tactical role (red: goalkeeper, blue: defenders, purple: midfielders, pink: forwards). Edges follow formation structure, connecting positionally adjacent roles within each team.
    }
    \label{fig:positional_graph}
\end{figure}

\noindent\textbf{Positional Roles.} Our proposed connectivity scheme respects tactical structure based on playing positions. Players are grouped by role (goalkeeper, defender, midfielder, forward), as shown in Figure \ref{fig:positional_graph}, using positional metadata from the tracking data. Within each team, edges connect adjacent tactical lines (goalkeeper to defenders, defenders to midfielders, midfielders to forwards), and the ball connects to all entities when present. This hierarchical connectivity encodes football-specific knowledge about team formations and plausible passing lanes.

\noindent\textbf{Missing Data.} Missing entities are assigned sentinel coordinates and excluded from graph message passing, allowing the model to leverage available spatial configurations and motion patterns even when ball tracking is unavailable.

\section{Experiments}
\subsection{Experimental Setup}
We benchmark various models on SoccerNet-GAR.

\noindent\textbf{Tracking-based models:} As a baseline tracking model, we employ a $20$-layer DeepGCN using GIN~\cite{xu2019powerfulgraphneuralnetworks} layers with sum aggregation, ReLU, residual connections, and layer normalization. Each node projects $8$-dimensional entity features to a $128$-dimensional embedding. Frame-level graph representations are obtained via mean pooling and temporally aggregated with MaxPool. Edges follow a positional role-based scheme connecting players by tactical roles (goalkeeper, defender, midfielder, forward). The classifier is a two-layer MLP (hidden dimension $256$, output dimension $10$), totaling $180$K parameters. Missing entities use sentinel coordinates ($-2.0$ after normalization) and are excluded from graph message passing.

\noindent\textbf{Video-based models:} The best video model is VideoMAEv2-Base~\cite{wang2023videomae}, a large transformer pretrained via masked autoencoding and finetuned end-to-end. The classifier is a two-layer MLP (hidden dimension $512$, output dimension $10$), totaling $86.3$M parameters.

\noindent\textbf{Training.} 
Tracking models are trained with Adam~\cite{kingma2014adam} (LR=$10^{-3}$), a ReduceLROnPlateau scheduler, and BS=$32$. 
For the video models, we use AdamW~\cite{loshchilov2017decoupled} with weight decay $10^{-4}$. Training configurations are defined based on whether the backbone is frozen or fine-tuned, with batch sizes chosen to satisfy memory constraints on a single V100 GPU.
Frozen backbones are trained with BS=$64$ and LR=$10^{-4}$. Fine-tuned models are trained with BS=$16$ and LR=$10^{-4}$, except for VideoMAEv2-B which uses BS=$8$ and LR=$5\times10^{-5}$.
Weighted random sampling draws $M=4{,}000$ samples per class per epoch to address class imbalance. Data augmentation includes team flip, spatial flips, and temporal jittering for tracking; horizontal flip and color jitter for video. Gradient clipping ($max\_norm=1.0$) stabilizes training.

\begin{table}[t]
\centering
\small
\begin{tabular}{lcccc}
\toprule
\textbf{Modality} & \textbf{Params} & \textbf{Bal. Acc.} & \textbf{F1} & \textbf{Training} \\
\midrule
Video & 86.3M & $60.9$ & $50.1$ & 28 GPU hours \\
Tracking & 180K & $\mathbf{77.8}$ & $\mathbf{57.0}$ & 4 GPU hours \\
\bottomrule
\end{tabular}
\caption{\textbf{Best Video vs. Baseline Tracking Model on SoccerNet-GAR.} Tracking outperforms video by $16.9$ percentage points in balanced accuracy and $6.9$ percentage points in macro F1 while having $479\times$ fewer parameters and $7\times$ faster training time.}
\label{tab:main_results}
\end{table}
\begin{figure*}[t]
    \centering
    \includegraphics[width=1.0\textwidth]{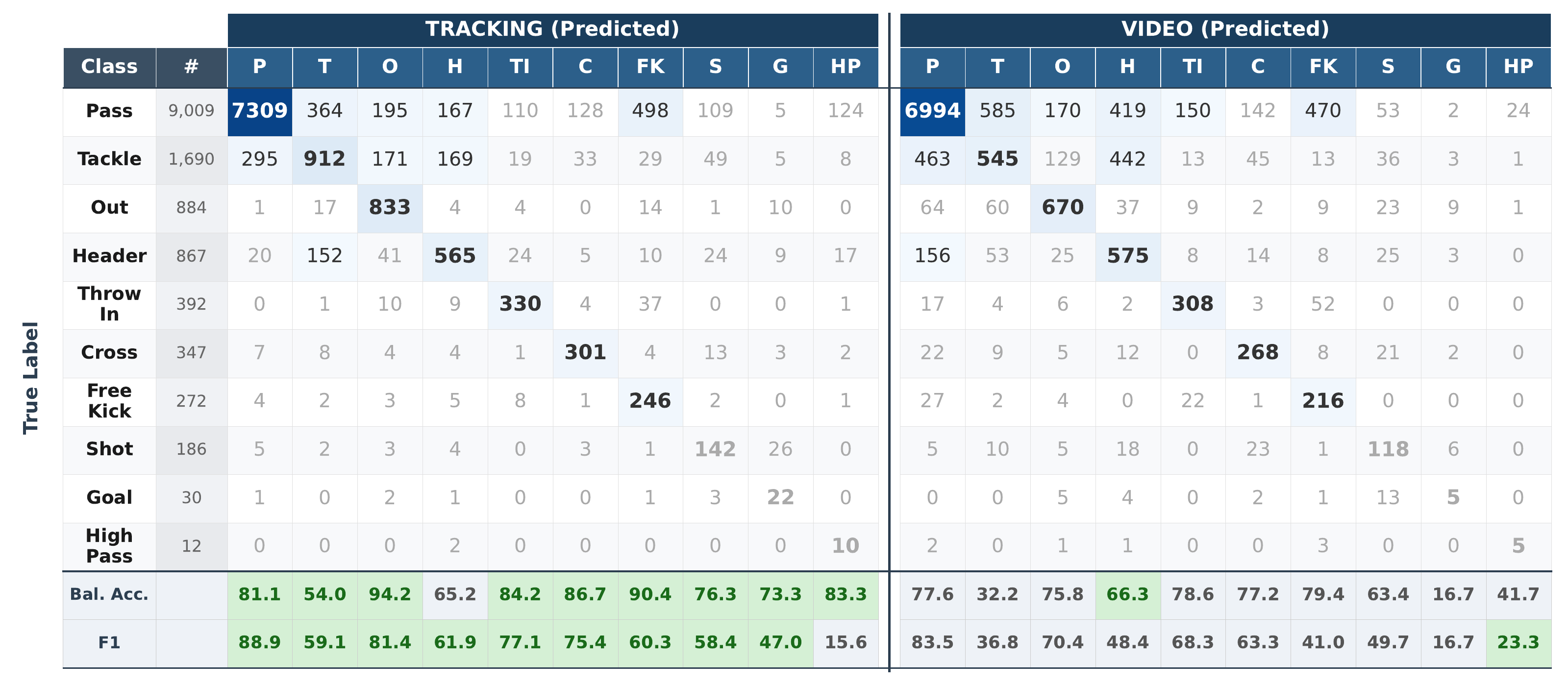}
    \caption{
    \textbf{Confusion Matrices and Per-class Metrics for our Baseline Tracking (GIN + MaxPool + Positional) and the Best Video (VideoMAEv2-B Finetuned) Models.}
    Green cells indicate the best modality per action class. Tracking dominates on $9$ of $10$ classes in balanced accuracy, with its largest gains on less occurring classes like \textit{GOAL} ($+56.7$ pp), and \textit{HIGH PASS} ($+41.7$ pp).
    }
    \label{fig:merged_cm}
\end{figure*}

\noindent\textbf{Metrics.} Following common practices, we report balanced accuracy (also corresponding to the average per-class recall) and macro F1 score as primary benchmarking metrics. 

\subsection{Main Results: Pixels vs Positions}
Table~\ref{tab:main_results} presents our primary comparison between the tracking and video modalities. Our baseline tracking model (GIN + Maxpool + Positional edges) achieves $77.8\%$ balanced accuracy and $57.0\%$ macro F1 using only $180$K parameters, training in approximately $4$ hours on a single V100 GPU. On the other hand, the best video baseline (VideoMAEv2-B finetuned) achieves $60.9\%$ balanced accuracy and $50.1\%$ macro F1 but requires $86.3$M parameters and $28$ GPU hours of training. The tracking model outperforms the video baseline by $16.9$ percentage points in balanced accuracy and $6.9$ percentage points in macro F1 while using $479\times$ fewer parameters and training with $7\times$ less GPU hours. This efficiency stems from the compact representation of tracking data, where each frame consists of $23$ nodes with $8$-dimensional features, compared to $224\times224\times3$ RGB tensors per video frame. All tracking results report mean $\pm$ standard deviation over $5$ random seeds; video results are from single run due to computational constraints.

\noindent\textbf{Per-class Analysis.} Figure~\ref{fig:merged_cm} compares per-class metrics for balanced accuracy (per-class recall) and macro F1 (per-class F1 score) across modalities. Tracking outperforms video on nine of ten classes, with particularly large margins on events that occur in characteristic pitch regions and induce structured player formations: \textit{GOAL} ($73.3\%$ vs $16.7\%$), \textit{HIGH PASS} ($83.3\%$ vs $41.7\%$), \textit{TACKLE} ($54.0\%$ vs $32.2\%$), and \textit{OUT} ($94.2\%$ vs $75.8\%$). Video shows a slight advantage only on \textit{HEADER} ($66.3\%$ vs $65.2\%$).
\textit{TACKLE} remains challenging for both modalities ($54.0\%$ tracking, $32.2\%$ video), as tackles involve rapid, localized interactions between two players that are difficult to distinguish from other close-contact situations in either representation.
Both modalities exhibit confusion between \textit{PASS} with \textit{FREE KICK} ($498$ and $470$ misclassifications of \textit{FREE KICK} as \textit{PASS} respectively for the tracking and video modality), likely because both involve a single player initiating ball distribution from a stationary position, producing similar spatial configurations despite the visual distinction of a set-piece setup.
Similarly, \textit{PASS} and \textit{TACKLE} as well as \textit{TACKLE} and \textit{HEADER} exhibit similar inter-class confusion for both modalities.
Finally, the results for \textit{GOAL} ($30$ test samples) and \textit{HIGH PASS} ($12$ test samples) demonstrate that tracking models can learn discriminative features even in severely data-scarce regimes ($73.3\%$ and $83.3\%$ respectively), whereas video models collapse on these classes ($16.7\%$ and $41.7\%$).

\subsection{Ablation Studies}
We evaluate design choices for both modalities. 
For tracking, we compare graph layer types, temporal aggregation strategies, and edge connectivity.
For video, we compare different backbone architectures, frozen and finetuned.

\begin{table}[t]
\centering
\small
\resizebox{1.0\linewidth}{!}{%
\begin{tabular}{lccccc}
\toprule
\textbf{Backbone} & \textbf{Frozen} & \textbf{Params} & \textbf{Bal. Acc.} & \textbf{F1} \\
\midrule
VideoMAE-B     & \checkmark & 86.3M & $34.6$ & $20.6$ \\
VideoMAE-B     & \ding{55}  & 86.3M & $55.2$ & $49.4$ \\
\midrule
VideoMAEv2-B   & \checkmark & 86.3M & $49.3$ & $30.7$ \\
\textbf{VideoMAEv2-B} & \ding{55} & \textbf{86.3M} & $\mathbf{60.9}$ & $\mathbf{50.1}$ \\
\bottomrule
\end{tabular}
}
\caption{\textbf{Video Backbone Ablation.} Finetuning yields substantial gains for both backbones, with VideoMAE showing the largest improvement ($+20.6\%$). VideoMAEv2-B achieves the best performance across both frozen and finetuned regimes.}
\label{tab:video_variants}
\end{table}

\noindent\textbf{Video Backbone Variants.} Table~\ref{tab:video_variants} compares two video-pretrained backbones, VideoMAE-B and VideoMAEv2-B, under two training regimes: frozen backbone with a trainable classification head versus full end-to-end finetuning.
Among frozen backbones, VideoMAEv2-B performs best ($49.3$\%), compared to VideoMAE-B ($34.6$\%). Finetuning substantially improves both models: VideoMAE-B gains $+20.6$ pp ($34.6$\% $\rightarrow$ $55.2$\%) and VideoMAEv2-B gains $+11.6$ pp ($49.3$\% $\rightarrow$ $60.9$\%). VideoMAEv2-B consistently achieves the best performance across both regimes, likely due to its larger and more diverse pretraining data compared to VideoMAE-B.

\noindent\textbf{Graph Layer Variants.} Table~\ref{tab:backbone_ablation} compares six graph convolutional operators with MaxPool temporal aggregation and positional edges in $20$-layer DeepGCN architectures with residual connections and layer normalization.
GIN achieves the highest balanced accuracy ($77.8 \pm 0.7\%$, $180$K parameters), very similar to GraphConv ($76.3 \pm 1.1\%$, $174$K). The near-identical performance suggests that positional edge structure provides most of the discriminative signal, with the choice of message-passing operator having minimal impact.
More expressive operators underperform: EdgeConv ($55.6 \pm 12.1\%$) and GATv2 ($61.8 \pm 5.4\%$) show both lower accuracy and higher variance. GEN ($72.8\%$) and GraphSAGE ($75.9\%$) are competitive but require, respectively, $1.9\times$ and $2.3\times$ more parameters. 
GIN and GraphSAGE also achieve the best F1 ($57.0$\% and $56.9\%$) with the lowest variance across both metrics for GIN, confirming its suitability as a strong tracking baseline.
\begin{table}[t]
\centering
\small
\begin{tabular}{lccc}
\toprule
\textbf{Backbone} & \textbf{Params} & \textbf{Bal. Acc.} & \textbf{F1} \\
\midrule
GraphConv & 174K & $76.3 \pm 1.1$ & $56.5 \pm 1.8$ \\
EdgeConv  & 174K & $55.6 \pm 12.1$ & $35.3 \pm 11.9$ \\
GATv2     & 177K & $61.8 \pm 5.4$ & $42.7 \pm 4.5$ \\
\textbf{GIN} & \textbf{180K} & $\mathbf{77.8 \pm 0.7}$ & $\mathbf{57.0 \pm 0.9}$ \\
GEN       & 341K & $72.8 \pm 3.0$ & $54.7 \pm 3.1$ \\
GraphSAGE & 421K & $75.9 \pm 1.1$ & $56.9 \pm 1.7$ \\
\bottomrule
\end{tabular}
\caption{\textbf{Graph Layer Ablation.} All models use temporal maxpool and positional edges. We report mean $\pm$ std over $5$ seeds. GIN achieves the highest balanced accuracy and macro F1 with the lowest variance across both metrics (bold). GraphConv and GraphSAGE perform comparably but with higher variance.
}
\label{tab:backbone_ablation}
\end{table}

\begin{table}[t]
\centering
\small
\resizebox{1.0\linewidth}{!}{%
\begin{tabular}{lcccc}
\toprule
\textbf{Edge Type} & \textbf{Params} & \textbf{Bal. Acc.} & \textbf{F1} \\
\midrule
No Edges & 180K & $68.9 \pm 2.7$ & $47.0 \pm 3.1$ \\
Fully Connected & 180K & $71.4 \pm 2.4$ & $49.8 \pm 1.6$ \\
Ball Distance ($r=20$m) & 180K & $68.6 \pm 1.0$ & $45.3 \pm 1.3$ \\
Ball KNN ($k=8$) & 180K & $67.0 \pm 0.9$ & $45.9 \pm 1.0$ \\
Distance ($r=15$m) & 180K & $68.0 \pm 1.1$ & $45.4 \pm 1.3$ \\
KNN ($k=8$) & 180K & $66.7 \pm 1.4$ & $46.4 \pm 1.6$ \\
\textbf{Positional} & \textbf{180K} & $\mathbf{77.8 \pm 0.7}$ & $\mathbf{57.0 \pm 0.9}$ \\
\bottomrule
\end{tabular}
}
\caption{\textbf{Edge Connectivity Ablation.}
All models use GIN backbone and temporal MaxPool. We report mean $\pm$ std over $5$ seeds.
Positional edges substantially outperform all alternatives in both metrics, leading the second-best edge type (Fully Connected) by $+6.4\%$ in balanced accuracy and $+7.2\%$ in F1.}
\label{tab:edge_ablation}
\end{table}
\begin{table}[t]
\centering
\small
\begin{tabular}{lcccc}
\toprule
\textbf{Temporal} & \textbf{Params} & \textbf{Bal. Acc.} & \textbf{F1} \\
\midrule
AvgPool   & 180K & $66.6 \pm 0.9$ & $42.5 \pm 1.5$ \\
MaxPool   & 180K & $\mathbf{77.8 \pm 0.7}$ & $57.0 \pm 0.9$ \\
Attention & 197K & $77.3 \pm 0.9$ & $58.3\pm 1.2$ \\
TCN       & 205K & $75.5 \pm 0.3$ & $\mathbf{58.4 \pm 0.6}$ \\
BiLSTM & 350K & $75.3 \pm 2.6$ & $57.9 \pm 1.0$ \\
\bottomrule
\end{tabular}
\caption{\textbf{Temporal Aggregation Ablation.}
All models use GIN backbone and positional edges and we report mean $\pm$ std over $5$ seeds. MaxPool achieves the highest balanced accuracy ($77.8\%$) while Attention and TCN achieve higher macro F1 ($58.3\%$ and $58.4\%$). The strong MaxPool performance indicates that simple temporal pooling remains competitive with learnable alternatives.
}
\label{tab:temporal_ablation}
\end{table}

\noindent\textbf{Edge Connectivity.} Table~\ref{tab:edge_ablation} compares seven connectivity patterns with GIN layers and MaxPool aggregation.
Positional edges achieve the best balanced accuracy ($77.8 \pm 0.7\%$) and F1 ($57 \pm 0.9\%$), with the lowest variance on both metrics. 
No-edges processing ($68.9 \pm 2.7\%$ balanced accuracy, $47.0 \pm 3.1\%$ F1) performs comparably to alternatives, confirming that message passing with poorly chosen edges provides little benefit over independent node processing. 
All geometric patterns cluster between $66.7$\%--$77.8$\% balanced accuracy, suggesting that unrestricted or proximity-based edges dilute the discriminative structure that positional connectivity preserves.

\noindent\textbf{Temporal Aggregation.} Table~\ref{tab:temporal_ablation} further extends our baseline model with more advanced temporal aggregation strategies still using GIN layers and positional edges over the $4.5$-second observation window at the cost of higher number of parameters.
MaxPool achieves the highest balanced accuracy ($77.8 \pm 0.7\%$, $180$K parameters), outperforming the learnable alternatives: Attention ($77.3 \pm 0.9\%$, $197$K), TCN ($75.5 \pm 0.3\%$, $205$K), and BiLSTM ($75.3 \pm 2.6\%$, $350$K), while AvgPool lags behind ($66.6 \pm 0.9\%$, $180$K).

However, TCN, Attention, and BiLSTM yield slightly higher macro F1 scores ($58.4\%$, $58.3\%$ and $57.9\%$ vs.\ $57.0\%$), suggesting that learnable temporal modules better balance per-class recall despite lower overall balanced accuracy.
We retain MaxPool as the default temporal aggregation for its simplicity, parameter efficiency, and strong balanced accuracy. Nonetheless, the competitive F1 gains from learnable modules indicate that the tracking modality backbone has room for improvement. Exploring more expressive temporal and spatio-temporal architectures intertwining the tracking backbone encoders with the temporal neck represents a promising direction for future work.
\section{Analysis}
\begin{figure}[t]
    \centering
    \includegraphics[width=\linewidth]{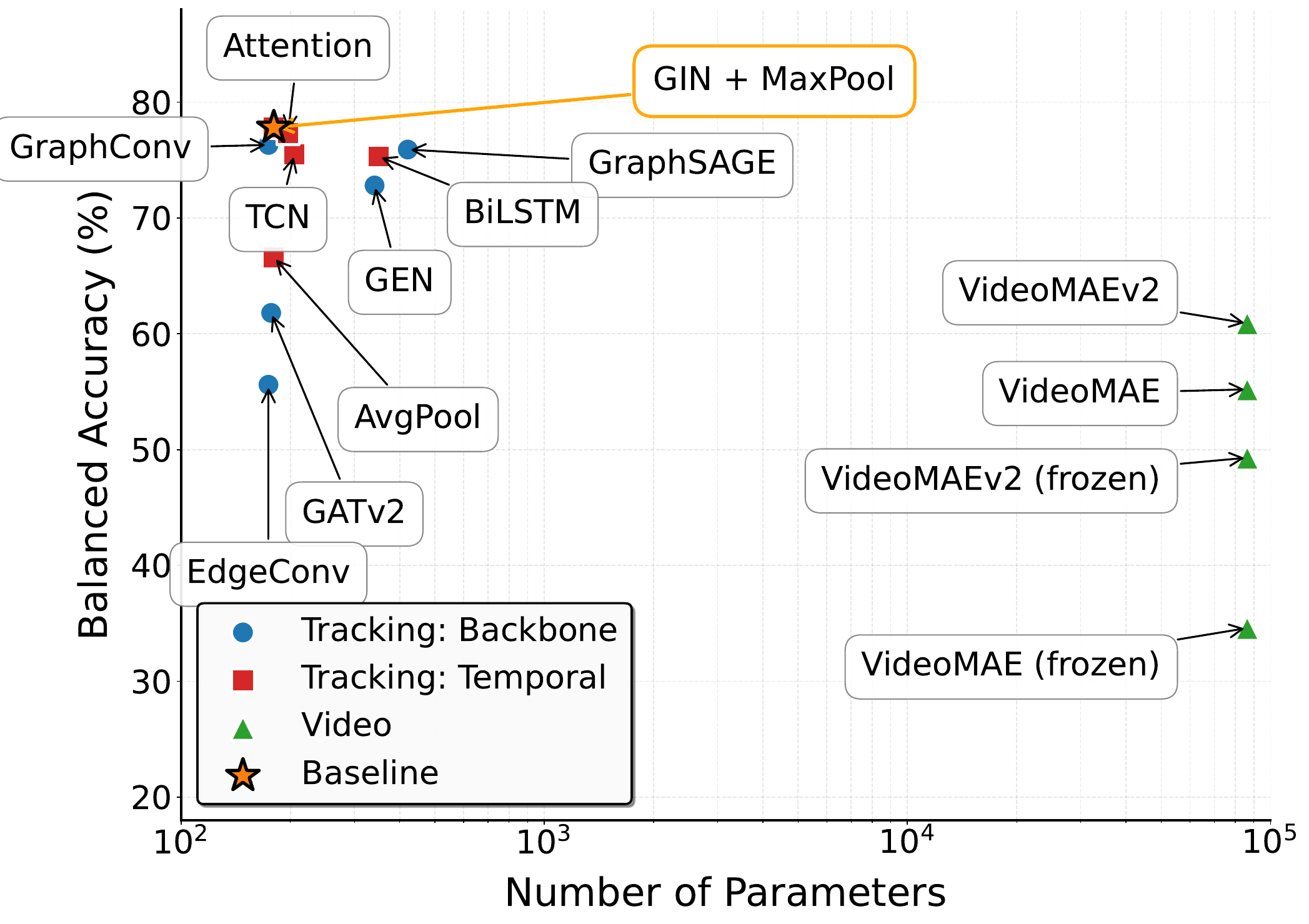}
    \caption{
    \textbf{Balanced Accuracy \vs Parameter Count.}
    Tracking models ($174$K--$421$K parameters) consistently outperform video models ($86.3$M parameters) in balanced accuracy despite using $200$-$496\times$ fewer parameters, highlighting that lightweight graph-based representations of player dynamics capture more discriminative signal for group activity recognition than large-scale video encoders.}
    \label{fig:params_vs_accuracy}
\end{figure}

\noindent\textbf{Model Parameters vs. Balanced Accuracy.}
Figure~\ref{fig:params_vs_accuracy} compares parameter efficiency across all evaluated models. Tracking models achieve $55.6$\%--$77.8$\% balanced accuracy with $174$K--$421$K parameters, while video models span $34.6$\%--$60.9$\% with $86.3$M parameters.
Tracking models exhibit clear clustering by component type. Backbone variants span $174$K--$421$K parameters with $55.6$\%--$77.8$\% accuracy, where GraphSAGE ($421$K) uses $2.3\times$ more parameters than GIN ($180$K) but underperforms by $1.9$\%. Temporal aggregation methods range from $180$K to $350$K parameters with $66.6$\%--$77.8$\% accuracy, where MaxPool ($180$K) achieves the highest accuracy at $77.8$\%, surpassing BiLSTM ($350$K,
$75.3$\%) despite using nearly half the parameters.
Video models cluster at $\sim$86.3M parameters with accuracy varying by $26.3$ percentage points ($34.6$\%--$60.9$\%). This large spread stems from pretraining strategy and frozen \vs finetuned regimes rather than architectural differences. VideoMAEv2-B frozen ($86.3$M, $49.3$\%) and VideoMAE frozen ($86.3$M, $34.6$\%) have identical parameter counts but differ by $14.7$ percentage points in accuracy, demonstrating that parameter count alone does not determine performance for video models.

\section{Conclusion}
We introduced SoccerNet-GAR, a novel dataset and benchmark for group activity recognition that aligns broadcast video with tracking data for $87{,}939$ football events across $10$ action classes. 
On this benchmark, our baseline tracking model reaches $77.8\%$ balanced accuracy and $57.0\%$ macro F1 with $180$K parameters, while the best video model reaches $60.9\%$ balanced accuracy and $50.1\%$ macro F1 with $86.3$M parameters. 
Tracking therefore improves performance by $16.9$ and $6.9$ percentage points, while using $479\times$ fewer parameters and training $7\times$ faster.
Our ablation studies demonstrate that domain-informed graph construction (positional edges encoding tactical structure) substantially outperforms geometric proximity patterns in both balanced accuracy and macro F1, and that exploring more expressive temporal architectures represent a promising direction for future work. Tracking excels on spatially distinctive events (\textit{OUT}, \textit{CROSS}, \textit{GOAL}) but slightly struggles with \textit{HEADER}, where video outperforms by $+1.1$ pp. Furthermore, tracking models can learn discriminative features even in severely data-scarce regimes, whereas video models collapse.
These findings indicate that explicit spatial relationships provide stronger signal than appearance cues for recognizing coordinated actions in sports. The complementary failure modes across modalities motivate future work on multimodal fusion, alongside dynamic role assignment, spatio-temporal backbone-neck modeling and causal (past-only) temporal windows for online deployment. We will release the dataset, code, and evaluation protocol.

\noindent\textbf{Acknowledgment.}
The research reported in this publication was supported by funding from King Abdullah University of Science and Technology (KAUST) - Center of Excellence for Generative AI, under award number 5940.
{
    \small
    \bibliographystyle{ieeenat_fullname}
    \bibliography{main}

@String(CVPR= {IEEE Conf. Comput. Vis. Pattern Recog.})

@String(ICCV= {Int. Conf. Comput. Vis.})

@String(ECCV= {Eur. Conf. Comput. Vis.})

@String(TOG= {ACM Trans. Graph.})

@String(ICLR = {Int. Conf. Learn. Represent.})

@String(AAAI = {AAAI})

@String(CVPRW= {IEEE Conf. Comput. Vis. Pattern Recog. Worksh.})

@String(CVPR  = {CVPR})

@String(ICCV  = {ICCV})

@String(ECCV  = {ECCV})

@String(TOG   = {ACM TOG})

@String(ICLR  = {ICLR})

@String(CVPRW= {CVPRW})

@inproceedings{bagautdinov2017socialscene,
  title={Social scene understanding: End-to-end multi-person action localization and collective activity recognition},
  author={Bagautdinov, Timur and Alahi, Alexandre and Fleuret, Fran{\c{c}}ois and Fua, Pascal and Savarese, Silvio},
  booktitle={CVPR},
  pages={4315--4324},
  year={2017}
}

@article{simeoni2025dinov3,
  title={Dinov3},
  author={Sim{\'e}oni, Oriane and Vo, Huy V and Seitzer, Maximilian and Baldassarre, Federico and Oquab, Maxime and Jose, Cijo and Khalidov, Vasil and Szafraniec, Marc and Yi, Seungeun and Ramamonjisoa, Micha{\"e}l and others},
  journal={arXiv preprint arXiv:2508.10104},
  year={2025}
}

@article{tong2022videomae,
  title={Videomae: Masked autoencoders are data-efficient learners for self-supervised video pre-training},
  author={Tong, Zhan and Song, Yibing and Wang, Jue and Wang, Limin},
  journal={Advances in neural information processing systems},
  volume={35},
  pages={10078--10093},
  year={2022}
}

@inproceedings{wang2023videomae,
  title={Videomae v2: Scaling video masked autoencoders with dual masking},
  author={Wang, Limin and Huang, Bingkun and Zhao, Zhiyu and Tong, Zhan and He, Yinan and Wang, Yi and Wang, Yali and Qiao, Yu},
  booktitle={Proceedings of the IEEE/CVF conference on computer vision and pattern recognition},
  pages={14549--14560},
  year={2023}
}

@inproceedings{radford2021learning,
  title={Learning transferable visual models from natural language supervision},
  author={Radford, Alec and Kim, Jong Wook and Hallacy, Chris and Ramesh, Aditya and Goh, Gabriel and Agarwal, Sandhini and Sastry, Girish and Askell, Amanda and Mishkin, Pamela and Clark, Jack and others},
  booktitle={International conference on machine learning},
  pages={8748--8763},
  year={2021},
  organization={PmLR}
}

@inproceedings{yan2018spatialtemporal,
  title={Spatial temporal graph convolutional networks for skeleton-based action recognition},
  author={Yan, Sijie and Xiong, Yuanjun and Lin, Dahua},
  booktitle={AAAI},
  volume={32},
  number={1},
  year={2018}
}

@inproceedings{wang2020learning,
  title={Learning visual context for group activity recognition},
  author={Yuan, Hangjie and Ni, Dong},
  booktitle={AAAI},
  volume={34},
  number={07},
  pages={3261--3269},
  year={2021}
}

@inproceedings{li2021groupformer,
  title={GroupFormer: Group activity recognition with clustered spatial-temporal transformer},
  author={Li, Shuaicheng and Cao, Qianggang and Liu, Lingbo and Yang, Kunlin and Liu, Shinan and Hou, Jun and Yi, Shuai},
  booktitle={ICCV},
  pages={13668--13677},
  year={2021}
}

@inproceedings{zhou2022composer,
  title={COMPOSER: Compositional reasoning of group activity in videos with keypoint-only modality},
  author={Zhou, Honglu and Kadav, Asim and Shamsian, Aviv and Geng, Shijie and Lai, Farley and Zhao, Long and Liu, Ting and Kapadia, Mubbasir and Graf, Hans Peter},
  booktitle={ECCV},
  pages={249--266},
  year={2022},
  organization={Springer}
}

@inproceedings{li2024skeleton,
  title={Skeleton-based group activity recognition via spatial-temporal panoramic graph},
  author={Li, Zhengcen and Chang, Xianxiang and Li, Yueran and Su, Jing},
  booktitle={ECCV},
  pages={254--270},
  year={2024},
  organization={Springer}
}

@inproceedings{choi2009they,
  title={What are they doing?: Collective activity classification using spatio-temporal relationship among people},
  author={Choi, Wongun and Shahid, Khuram and Savarese, Silvio},
  booktitle={2009 IEEE 12th international conference on computer vision workshops, ICCV Workshops},
  pages={1282--1289},
  year={2009},
  organization={IEEE}
}

@inproceedings{ibrahim2016hierarchical,
  title={A hierarchical deep temporal model for group activity recognition},
  author={Ibrahim, Mostafa S and Muralidharan, Srikanth and Deng, Zhiwei and Vahdat, Arash and Mori, Greg},
  booktitle={Proceedings of the IEEE conference on computer vision and pattern recognition},
  pages={1971--1980},
  year={2016}
}

@misc{yan2020socialadaptivemoduleweaklysupervised,
  title={Social Adaptive Module for Weakly-supervised Group Activity Recognition}, 
  author={Rui Yan and Lingxi Xie and Jinhui Tang and Xiangbo Shu and Qi Tian},
  year={2020},
  eprint={2007.09470},
  archivePrefix={arXiv},
  primaryClass={cs.CV}
}

@inproceedings{deliege2021soccernet,
  title={SoccerNet-v2: A dataset and benchmarks for holistic understanding of broadcast soccer videos},
  author={Deli{\`e}ge, Adrien and Cioppa, Anthony and Giancola, Silvio and Seikavandi, Meisam J and Dueholm, Jacob V and Nasrollahi, Kamal and Ghanem, Bernard and Moeslund, Thomas B and Van Droogenbroeck, Marc},
  booktitle={CVPR Workshops},
  year={2021}
}

@article{kim2023towards,
  title={Towards more practical group activity detection: A new benchmark and model},
  author={Kim, Dongkeun and Song, Youngkil and Cho, Minsu and Kwak, Suha},
  journal={arXiv preprint arXiv:2312.02878},
  year={2023}
}

@article{pei2024fifawc,
  title={FIFAWC: a dataset with detailed annotation and rich semantics for group activity recognition},
  author={Pei, Duoxuan and Huang, Di and Wang, Yunhong},
  journal={Frontiers of Computer Science},
  volume={18},
  number={6},
  pages={186351},
  year={2024}
}

@article{cioppa2022scaling,
  title={Scaling up SoccerNet with multi-view spatial localization and re-identification},
  author={Cioppa, Anthony and Deliege, Adrien and Giancola, Silvio and Ghanem, Bernard and Van Droogenbroeck, Marc},
  journal={Scientific data},
  volume={9},
  number={1},
  pages={355},
  year={2022},
  publisher={Nature Publishing Group UK London}
}

@inproceedings{giancola2018soccernet,
  title={{SoccerNet}: A Scalable Dataset for Action Spotting in Soccer Videos},
  author={Giancola, Silvio and Amine, Mohieddine and Dghaily, Tarek and Ghanem, Bernard},
  booktitle={IEEE Conference on Computer Vision and Pattern Recognition Workshops (CVPRW)},
  pages={1711--1721},
  year={2018}
}

@article{cioppa2024soccernet,
  title={{SoccerNet} 2023 Challenges Results},
  author={Cioppa, Anthony and Giancola, Silvio and Somers, Vladimir and Magera, Floriane and Zhou, Xin and Mkhallati, Hassan and Deli{\`e}ge, Adrien and Held, Jan and Hinojosa, Carlos and Mansourian, Amir M and others},
  journal={Sports Engineering},
  volume={27},
  number={2},
  pages={24},
  year={2024}
}

@inproceedings{somers2024soccernet,
  title={{SoccerNet} Game State Reconstruction: End-to-End Athlete Tracking and Identification on a Minimap},
  author={Somers, Vladimir and Joos, Victor and Cioppa, Anthony and Giancola, Silvio and Ghasemzadeh, Seyed Abolfazl and Magera, Floriane and Standaert, Baptiste and Mansourian, Amir Mohammad and Zhou, Xin and Kasaei, Shohreh and others},
  booktitle={IEEE Conference on Computer Vision and Pattern Recognition Workshops (CVPRW)},
  year={2024}
}

@inproceedings{yeh2019forecasting,
  title={Forecasting Basketball Trajectories and Player Intentions using Graph Neural Networks},
  author={Yeh, Raymond A and Schwing, Alexander G and Huang, Jonathan and Murphy, Kevin},
  booktitle={ICML Workshop on Computer Vision for Autonomous Vehicles},
  year={2019}
}

@inproceedings{fernandez2019wide,
  title={Wide Open Spaces: A Statistical Technique for Measuring Space Creation in Professional Soccer},
  author={Fern{\'a}ndez, Javier and Bornn, Luke and Cervone, Dan},
  booktitle={MIT Sloan Sports Analytics Conference},
  year={2019}
}

@inproceedings{cartas2022graphbased,
  title={A Graph-Based Method for Soccer Action Spotting Using Unsupervised Player Classification},
  author={Cartas, Alejandro and Ballester, Coloma and Haro, Gloria},
  booktitle={ACM International Workshop on Multimedia Content Analysis in Sports (MMSports)},
  pages={93--102},
  year={2022}
}

@article{majeed2025realtime,
  title={Real-time Analysis of Soccer Ball-Player Interactions Using Graph Convolutional Networks for Enhanced Game Insights},
  author={Majeed, Abdul and Hashmi, Mohammad Farukh and Ashraf, Muhammad Umar and Srivastava, Gitanjali and Geem, Zong Woo and Bokde, Neeraj Dhanraj},
  journal={Scientific Reports},
  volume={15},
  number={1},
  pages={1--19},
  year={2025}
}

@inproceedings{kipf2016gcn,
  title={Semi-Supervised Classification with Graph Convolutional Networks},
  author={Kipf, Thomas N and Welling, Max},
  booktitle={International Conference on Learning Representations (ICLR)},
  year={2017}
}

@inproceedings{hamilton2017graphsage,
  title={Inductive Representation Learning on Large Graphs},
  author={Hamilton, Will and Ying, Zhitao and Leskovec, Jure},
  booktitle={Advances in Neural Information Processing Systems (NeurIPS)},
  pages={1024--1034},
  year={2017}
}

@inproceedings{fang2018learning,
  title={Learning pose grammar to encode human body configuration for 3d pose estimation},
  author={Fang, Hao-Shu and Xu, Yuanlu and Wang, Wenguan and Liu, Xiaobai and Zhu, Song-Chun},
  booktitle={Proceedings of the AAAI conference on artificial intelligence},
  volume={32},
  number={1},
  year={2018}
}

@inproceedings{wang2019dynamicedgeconv,
  title={Dynamic Graph {CNN} for Learning on Point Clouds},
  author={Wang, Yue and Sun, Yongbin and Liu, Ziwei and Sarma, Sanjay E and Bronstein, Michael M and Solomon, Justin M},
  booktitle={ACM Transactions on Graphics (TOG)},
  volume={38},
  number={5},
  pages={1--12},
  year={2019}
}

@inproceedings{li2019deepgcns,
  title={{DeepGCNs}: Can {GCNs} Go As Deep As {CNNs}?},
  author={Li, Guohao and Muller, Matthias and Thabet, Ali and Ghanem, Bernard},
  booktitle={IEEE International Conference on Computer Vision (ICCV)},
  pages={9267--9276},
  year={2019}
}

@inproceedings{lea2017temporalconvolutional,
  title={Temporal Convolutional Networks for Action Segmentation and Detection},
  author={Lea, Colin and Flynn, Michael D and Vidal, Rene and Reiter, Austin and Hager, Gregory D},
  booktitle={IEEE Conference on Computer Vision and Pattern Recognition (CVPR)},
  pages={156--165},
  year={2017}
}

@inproceedings{gavrilyuk2020actor,
  title={Actor-Transformers for Group Activity Recognition},
  author={Gavrilyuk, Kirill and Sanford, Ryan and Javan, Mehrsan and Snoek, Cees GM},
  booktitle={IEEE Conference on Computer Vision and Pattern Recognition (CVPR)},
  pages={839--848},
  year={2020}
}

@article{ding2020graphattention,
  title={A Graph Attention Based Approach for Trajectory Prediction in Multi-Agent Sports Games},
  author={Ding, Dawei and Huang, Hsiangsheng},
  journal={arXiv preprint arXiv:2012.10531},
  year={2020}
}

@misc{pff2022worldcupdataset,
    author = {{PFF FC}},
    title = {Data-Driven Exploration of the 2022 FIFA World Cup},
    howpublished = {\url{https://www.blog.fc.pff.com/blog/enhanced-2022-world-cup-dataset}},
    year = {2023},
    note = {Accessed: November 12, 2025}
}

@article{kingma2014adam,
  title={Adam: A method for stochastic optimization},
  author={Kingma, Diederik P},
  journal={arXiv preprint arXiv:1412.6980},
  year={2014}
}

@article{loshchilov2017decoupled,
  title={Decoupled weight decay regularization},
  author={Loshchilov, Ilya and Hutter, Frank},
  journal={arXiv preprint arXiv:1711.05101},
  year={2017}
}

@misc{benzakour2024oslactionspottingunifiedlibraryaction,
      title={OSL-ActionSpotting: A Unified Library for Action Spotting in Sports Videos}, 
      author={Yassine Benzakour and Bruno Cabado and Silvio Giancola and Anthony Cioppa and Bernard Ghanem and Marc Van Droogenbroeck},
      year={2024},
      eprint={2407.01265},
      archivePrefix={arXiv},
      primaryClass={cs.CV},
      url={https://arxiv.org/abs/2407.01265}, 
}

@misc{xu2019powerfulgraphneuralnetworks,
      title={How Powerful are Graph Neural Networks?}, 
      author={Keyulu Xu and Weihua Hu and Jure Leskovec and Stefanie Jegelka},
      year={2019},
      eprint={1810.00826},
      archivePrefix={arXiv},
      primaryClass={cs.LG},
      url={https://arxiv.org/abs/1810.00826}, 
}

@inproceedings{Cioppa2021Camera,
	title = {Camera Calibration and Player Localization in {SoccerNet-v2} and Investigation of their Representations for Action Spotting},
	author = {Cioppa, Anthony and Deli{\`e}ge, Adrien and Giancola, Silvio and Magera, Floriane and Barnich, Olivier and Ghanem, Bernard and Van Droogenbroeck, Marc},
	booktitle = cvsports,
	month = Jun,
	pages = {4532-4541},
	year = {2021},
	address = nashville,
	url = {http://hdl.handle.net/2268/259026},
	doi = {10.1109/CVPRW53098.2021.00511}
}

@misc{brody2022attentivegraphattentionnetworks,
      title={How Attentive are Graph Attention Networks?}, 
      author={Shaked Brody and Uri Alon and Eran Yahav},
      year={2022},
      eprint={2105.14491},
      archivePrefix={arXiv},
      primaryClass={cs.LG},
      url={https://arxiv.org/abs/2105.14491}, 
}

@article{graves2005framewise,
  title={Framewise phoneme classification with bidirectional LSTM and other neural network architectures},
  author={Graves, Alex and Schmidhuber, J{\"u}rgen},
  journal={Neural networks},
  volume={18},
  number={5-6},
  pages={602--610},
  year={2005},
  publisher={Elsevier}
}

@article{Leduc2024SoccerNetDepth,
  title={SoccerNet-Depth: a Scalable Dataset for Monocular Depth Estimation in Sports Videos},
  author={Arnaud Leduc and Anthony Cioppa and Silvio Giancola and Bernard Ghanem and Marc Van Droogenbroeck},
  journal={2024 IEEE/CVF Conference on Computer Vision and Pattern Recognition Workshops (CVPRW)},
  year={2024},
  pages={3280-3282},
  url={https://api.semanticscholar.org/CorpusID:272703273}
}

@article{Gautam2024SoccerNetEchoes,
  title={SoccerNet-Echoes: A Soccer Game Audio Commentary Dataset},
  author={Sushant Gautam and Mehdi Houshmand Sarkhoosh and Jan Held and Cise Midoglu and Anthony Cioppa and Silvio Giancola and Vajira Thambawita and Michael Alexander Riegler and P. Halvorsen and Mubarak Shah},
  journal={2024 International Symposium on Multimedia (ISM)},
  year={2024},
  pages={71-78},
  url={https://api.semanticscholar.org/CorpusID:269757092}
}

@article{Sarkhoosh2025Beyond,
  title={Beyond Audio: Enhancing SoccerNet-Echoes with Multimodal Event Extraction Using LLMs},
  author={Mehdi Houshmand Sarkhoosh and Sushant Gautam and Cise Midoglu and Thu Nguyen and Jan Held and Anthony Cioppa and Silvio Giancola and Vajira Thambawita and Michael A. Riegler and P{\aa}l Halvorsen},
  journal={Int. J. Semantic Comput.},
  year={2025},
  volume={19},
  pages={589-613},
  url={https://api.semanticscholar.org/CorpusID:282442247}
}

@article{Cioppa2022SoccerNetTracking,
  title={SoccerNet-Tracking: Multiple Object Tracking Dataset and Benchmark in Soccer Videos},
  author={Anthony Cioppa and Silvio Giancola and Adrien Deli{\`e}ge and Le Kang and Xin Zhou and Zhiyu Cheng and Bernard Ghanem and Marc Van Droogenbroeck},
  journal={2022 IEEE/CVF Conference on Computer Vision and Pattern Recognition Workshops (CVPRW)},
  year={2022},
  pages={3490-3501},
  url={https://api.semanticscholar.org/CorpusID:248177722}
}

@article{Istasse2023DeepSportradarv2,
  title={DeepSportradar-v2: A Multi-Sport Computer Vision Dataset for Sport Understandings},
  author={Maxime Istasse and Vladimir Somers and Pratheeban Elancheliyan and Jaydeep De and Davide Zambrano},
  journal={Proceedings of the 6th International Workshop on Multimedia Content Analysis in Sports},
  year={2023},
  url={https://api.semanticscholar.org/CorpusID:264450218}
}

@article{Giancola2024Deep,
  title={Deep learning for action spotting in association football videos},
  author={Silvio Giancola and Anthony Cioppa and Bernard Ghanem and Marc Van Droogenbroeck},
  journal={ArXiv},
  year={2024},
  volume={abs/2410.01304},
  url={https://api.semanticscholar.org/CorpusID:273025740}
}

@article{Host2022AnOverview,
  title={An overview of Human Action Recognition in sports based on Computer Vision},
  author={Kristina Host and Marina Ivasic-Kos},
  journal={Heliyon},
  year={2022},
  volume={8},
  url={https://api.semanticscholar.org/CorpusID:249421447}
}

@inproceedings{stockl2021making,
  title={Making offensive play predictable-using a graph convolutional network to understand defensive performance in soccer},
  author={St{\"o}ckl, Michael and Seidl, Thomas and Marley, Daniel and Power, Paul},
  booktitle={Proceedings of the 15th MIT sloan sports analytics conference},
  volume={2022},
  year={2021},
  organization={MIT sloan sports analytics conference United States}
}

@misc{wang2023tacticaiaiassistantfootball,
      title={TacticAI: an AI assistant for football tactics}, 
      author={Zhe Wang and Petar Veličković and Daniel Hennes and Nenad Tomašev and Laurel Prince and Michael Kaisers and Yoram Bachrach and Romuald Elie and Li Kevin Wenliang and Federico Piccinini and William Spearman and Ian Graham and Jerome Connor and Yi Yang and Adrià Recasens and Mina Khan and Nathalie Beauguerlange and Pablo Sprechmann and Pol Moreno and Nicolas Heess and Michael Bowling and Demis Hassabis and Karl Tuyls},
      year={2023},
      eprint={2310.10553},
      archivePrefix={arXiv},
      primaryClass={cs.LG},
      url={https://arxiv.org/abs/2310.10553}, 
}

@inproceedings{yeh2019diverse,
  title={Diverse generation for multi-agent sports games},
  author={Yeh, Raymond A and Schwing, Alexander G and Huang, Jonathan and Murphy, Kevin},
  booktitle={Proceedings of the IEEE/CVF Conference on Computer Vision and Pattern Recognition},
  pages={4610--4619},
  year={2019}
}

@inproceedings{anzer2022detection,
  title={Detection of tactical patterns using semi-supervised graph neural networks},
  author={Anzer, Gabriel and Bauer, Pascal and Brefeld, Ulf and Fa{\ss}meyer, Dennis},
  booktitle={16th MIT sloan sports analytics conference},
  pages={1--15},
  year={2022}
}

@article{hauri2022group,
  title={Group activity recognition in basketball tracking data--neural embeddings in team sports (NETS)},
  author={Hauri, Sandro and Vucetic, Slobodan},
  journal={arXiv preprint arXiv:2209.00451},
  year={2022}
}

@inproceedings{zhang2024bicausal,
  title={Bi-Causal: Group Activity Recognition via Bidirectional Causality},
  author={Zhang, Youliang and Liu, Wenxuan and Xu, Danni and Zhou, Zhuo and Wang, Zheng},
  booktitle={IEEE/CVF Conference on Computer Vision and Pattern Recognition (CVPR)},
  year={2024}
}

@article{xie2024active,
  title={Active Factor Graph Network for Group Activity Recognition},
  author={Xie, Zhao and Jiao, Chang and Wu, Kewei and Guo, Dan and Hong, Richang},
  journal={IEEE Transactions on Image Processing},
  volume={33},
  pages={1574--1587},
  year={2024}
}
}

\clearpage

\twocolumn[{
\centering
\includegraphics[width=\textwidth]{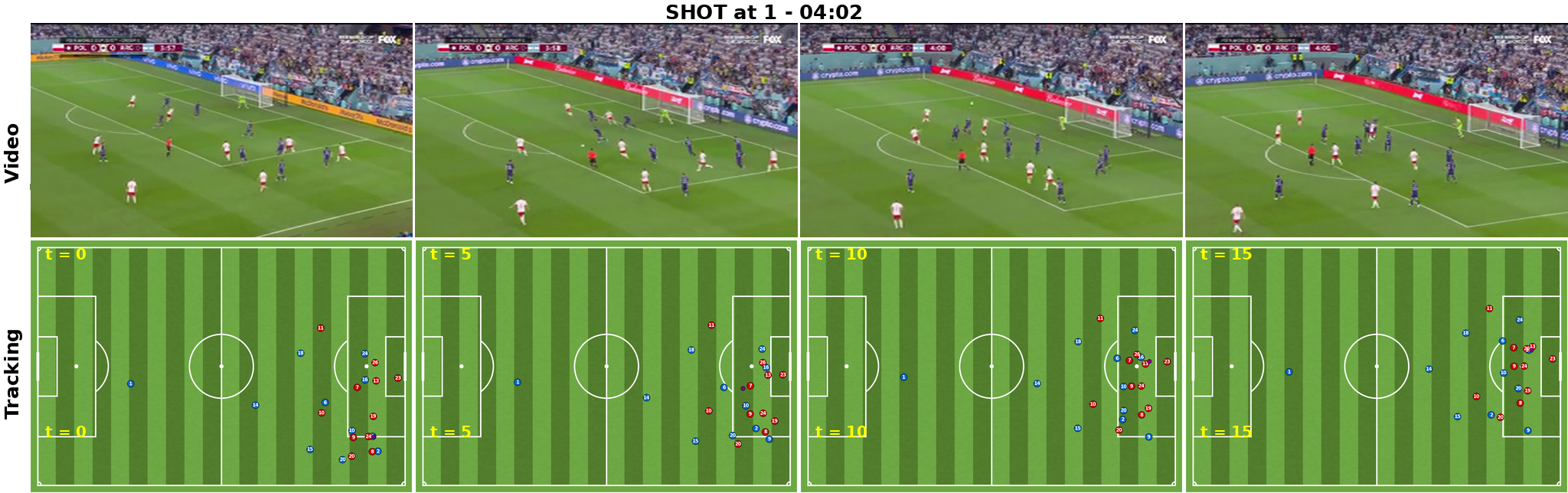}
\captionof{figure}{
\textbf{Video vs Tracking Input for a SHOT Event.}
Top: broadcast frames at four timesteps showing camera motion and partial field coverage. Bottom: synchronized tracking data on a standardized pitch, providing complete spatial context for all players regardless of camera position.
}
\label{fig:modality_comparison}
\vspace{0.5cm}
}]

\section{Visualization of data synchronization}
Figure~\ref{fig:modality_comparison} illustrates the two input modalities for a \textit{SHOT} event at match time 04:02. The top row shows four broadcast frames sampled at timesteps $t \in \{0, 5, 10, 15\}$ within the observation window. The camera pans and zooms to follow the play, introducing viewpoint changes, partial occlusions, and background clutter that the video encoder must contend with. The bottom row shows the corresponding tracking data projected onto a standardized pitch template, where red and blue markers denote players from opposing teams and the purple marker indicates the ball position. The tracking representation abstracts away visual noise and captures the spatial configuration of all $22$ players in a canonical coordinate frame, making tactical patterns such as defensive compactness and attacking movement directly readable. This example highlights a key advantage of tracking modality: while the broadcast view loses several players outside the frame at each timestep, the bird's-eye tracking view preserves the full team-level spatial context throughout the observation window.

\begin{figure}[!t]
\centering
\begin{tikzpicture}
\begin{axis}[
    xlabel={Number of Training Matches},
    ylabel={Balanced Accuracy (\%)},
    grid=major,
    width=1.0\columnwidth,
    height=0.7\columnwidth,
    xmin=0, xmax=50,
    ymin=35, ymax=85,
    legend pos=south east,
]
\addplot[mark=*, blue, thick] coordinates {
    (5, 67.0)
    (10, 71.0)
    (15, 75.3)
    (20, 75.0)
    (25, 77.0)
    (30, 77.3)
    (35, 78.5)
    (40, 78.4)
    (45, 77.8)
};
\addlegendentry{Tracking}

\addplot[mark=*, red, thick] coordinates {
    (5, 41.6)
    (10, 48.8)
    (15, 49.1)
    (20, 51.6)
    (25, 55.4)
    (30, 55.9)
    (35, 62.4)
    (40, 59.8)
    (45, 60.9)
};
\addlegendentry{Video}
\end{axis}
\end{tikzpicture}
\caption{
\textbf{Performance \vs Number of Training Matches for Video and Tracking Modalities.}
Both modalities plateau around $\approx 35$ matches, but tracking reaches strong performance ($67.0$\%) with as few as $5$ matches, whereas video starts at $41.6$\%. The gap narrows from $25.4$\% ($5$ matches) to $16.1$\% ($35$ matches).}
\label{fig:training_matches}
\end{figure}

\section{Data scaling}
Figure~\ref{fig:training_matches} shows performance versus training set size for both modalities, using GIN + MaxPool + Positional edges for tracking and VideoMAEv2-B finetuned for video (mean over $5$ seeds for tracking). Tracking achieves $67.0$\% with only $5$ matches and plateaus around $35$ matches ($78.5\%$), while video achieves $41.6$\% with $5$ matches and also plateaus around $35$ matches ($62.4$\%), after which both modalities fluctuate slightly. The gap narrows from $25.4$\% at $5$ matches to $16.1$\% at $35$ matches, indicating that video benefits more from additional data but remains substantially behind. Tracking maintains superior performance across all data regimes, suggesting that structured positional representations require less training data to reach strong performance.

\mbox{}
\clearpage

\end{document}